%% file: paper.tex
\documentclass{article} 
\usepackage{nips14submit_e,times}
\usepackage[hidelinks]{hyperref}
\usepackage{url}

\usepackage{cite}
\usepackage[pdftex]{graphicx}
\usepackage{wrapfig}
\usepackage{algorithm}
\usepackage{algorithmicx}
\usepackage{algpseudocode}
\usepackage{subcaption}
\usepackage{multirow}
\usepackage{mdwmath}
\usepackage{mdwtab}
\usepackage{amssymb}
\usepackage{amsmath}
\usepackage[font=footnotesize]{caption}
\usepackage{tabularx}
\usepackage{mdwlist}
\newcommand{\ie}{\emph{i.e.}}
\newcommand{\etal}{\emph{et~al.}}

\newcommand{\comment}[1]{}

\title{Multimodal Sparse Coding for Event Detection}

\author{
Youngjune Gwon\quad William M. Campbell\quad Kevin Brady\quad Douglas Sturim\\MIT Lincoln Laboratory, Lexington, MA 02420, USA
\And Miriam Cha\quad H. T. Kung\\Harvard University, Cambridge, MA 02138, USA
}

\nipsfinalcopy 

\begin{document}
\footnotetext{This work was sponsored by the Department of Defense under Air Force Contract FA8721-05-C-0002. Opinions, interpretations, conclusions, and recommendations are those of the authors and are not necessarily endorsed by the United States Government.}
\maketitle

\begin{abstract}
Unsupervised feature learning methods have proven effective for classification tasks based on a single modality. We present multimodal sparse coding for learning feature representations shared across multiple modalities. The shared representations are applied to multimedia event detection (MED) and evaluated in comparison to unimodal counterparts, as well as other feature learning methods such as GMM supervectors and sparse RBM. We report the cross-validated classification accuracy and mean average precision of the MED system trained on features learned from our unimodal and multimodal settings for a subset of the TRECVID MED 2014 dataset.  
\end{abstract}

\input{intro}
\input{avf}
\input{eval}

\input{conc}

\small
{
\bibliographystyle{abbrv}
\bibliography{paper}
}
\end{document}

%% file: intro.tex
\section{Introduction}
Multimedia Event Detection (MED) aims to identify complex activities occurring at a specific place and time involving various interactions of human actions and objects. MED is considered more difficult than concept analysis such as action recognition and has received significant attention in computer vision and machine learning research. In this paper, we propose the use of sparse coding for multimodal feature learning in the context of MED. Originally proposed to explain neurons encoding sensory information \cite{olshausen}, sparse coding provides an unsupervised method to learn basis vectors for efficient data representation. More recently, sparse coding has been used to model the relationship between correlated data sources. By jointly training dictionaries with audio and video tracks from the same multimedia clip, we can force the two modalities to share a similar sparse representation whose benefit includes robust detection and cross-modality retrieval. 

In the next section, we will describe audio-video feature learning in various unimodal and multimodal settings for sparse coding. We then present our experiments with TRECVID MED dataset. We will discuss the empirical results, compare them to other methods, and conclude. 

%% file: avf.tex
\section{Audio-video Feature Learning}
In summary, our approach is to build feature vectors by sparse coding on the low-level audio and video features. Multiple feature vectors (\ie, sparse codes) are aggregated via max pooling. The resulting pooled feature vectors can scale to file level, and we use them to train an array of classifiers for MED. 

\subsection{Low-level feature extraction and preprocessing}
We begin by locating the keyframes of a given multimedia clip. We apply a simple two-pass algorithm that computes color histogram difference of any two successive frames and determines a keyframe candidate based on the threshold calculated on the mean and standard deviation of the histogram differences. We examine the number of different colors present in the keyframe candidates and discard the ones with less than 26 colors. This ensures that our keyframes are not all-black or all-white blank images.

\begin{figure}
\centering
\begin{subfigure}[b]{0.4\textwidth}
\centering
\includegraphics[width=1\textwidth]{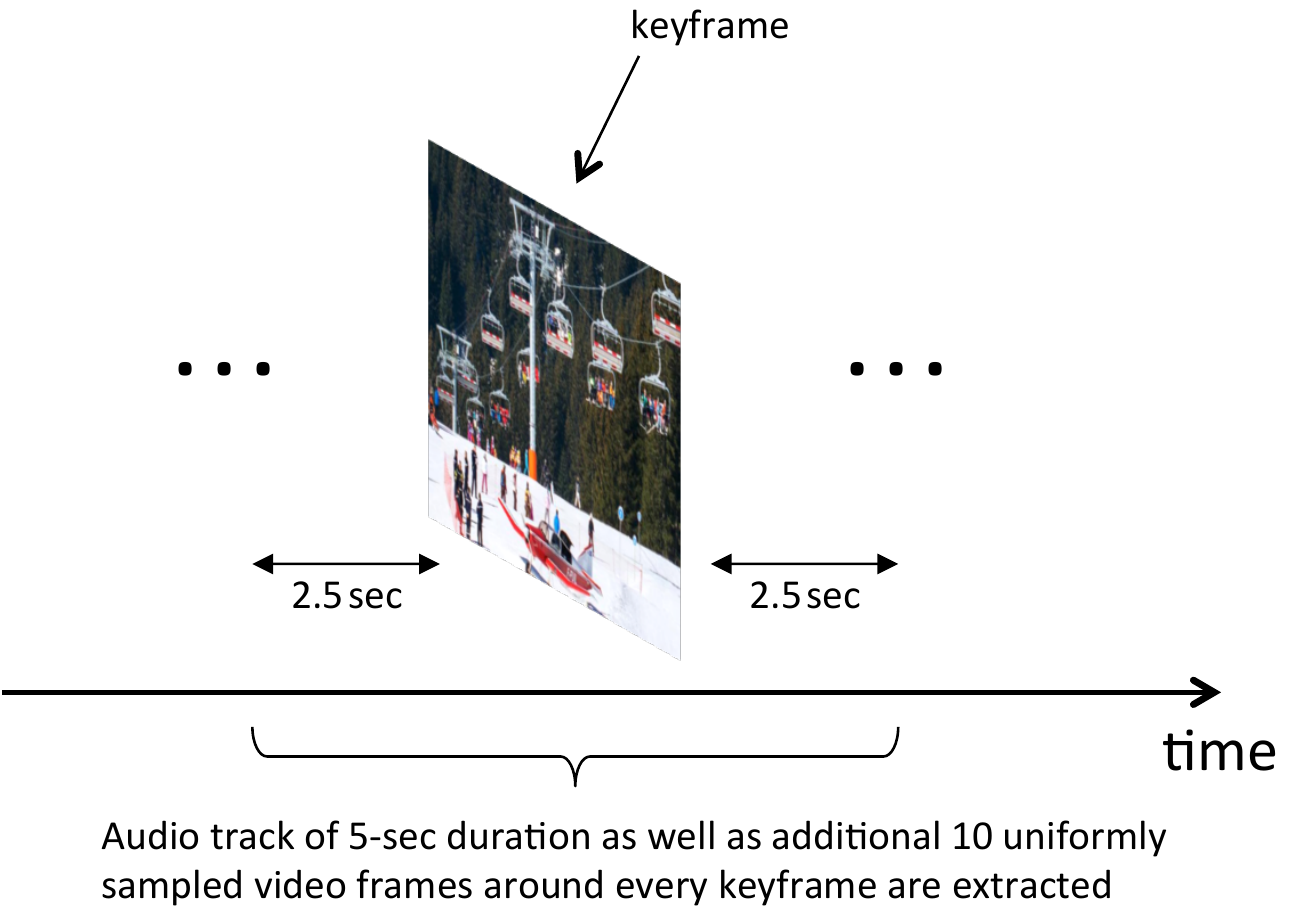}
\caption{Keyframe extraction}
\label{subfig:keyframe-ext}
\end{subfigure}
~
\begin{subfigure}[b]{0.25\textwidth}
\centering
\includegraphics[width=0.5\textwidth]{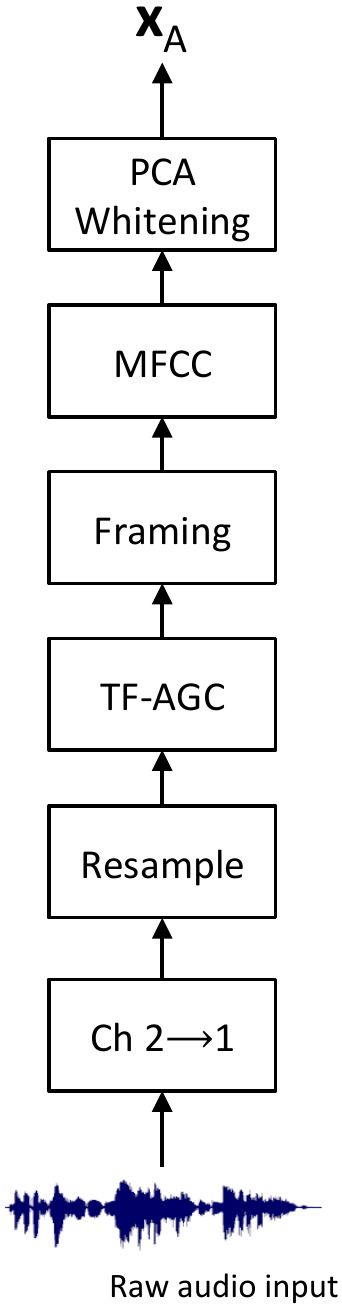}
\caption{Audio preprocessing}
\label{subfig:audio-preproc}
\end{subfigure}
~
\begin{subfigure}[b]{0.25\textwidth}
\centering
\includegraphics[width=0.34\textwidth]{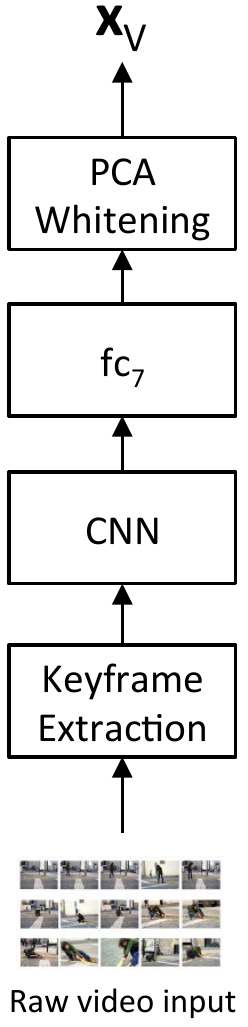}
\caption{Video preprocessing}
\label{subfig:video-preproc}
\end{subfigure}
\caption{Preprocessing audio and video data from multimedia clip}
\label{fig:preproc}
\end{figure}

Around each keyframe, we extract 5-sec audio data and additional 10 uniformly sampled video frames within the duration as illustrated in Figure~\ref{subfig:keyframe-ext}. If extracted audio is stereo, we take only the left channel. The audio waveform is resampled to 22.05\,kHz and regularized by the time-frequency automatic gain control (TF-AGC) to balance the energy in sub-bands. We form audio frames using a 46-msec Hann window with 50\% overlap between successive frames for smoothing. For each frame, we compute 16 the Mel-frequency cepstral coefficients (MFCCs) as the low-level audio feature. In addition, we append 16 delta cepstral and 16 delta-delta cepstral coefficients, which make our low-level audio feature vectors 48 dimensional. Finally, we apply PCA whitening before unsupervised learning. The complete audio preprocessing steps are described in Figure~\ref{subfig:audio-preproc}.

For video preprocessing, we take a deep learning approach. We have tried out pretrained convolutional neural network (CNN) models and ended up choosing \texttt{VGG\_ILSVRC\_19\_layers}, the 19-layer model by University of Oxford's Visual Geometry Group (VGG) \cite{vgg} for the ImageNet Large-scale Visual Recognition Challenge (ILSVRC). As depicted in Figure~\ref{subfig:video-preproc}, we run the CNN feedforward passes with the extracted video frames. For each video frame, we take 4,096-dimensional hidden activation from fc$_7$, the highest hidden layer before the final ReLU (\ie, the rectification non-linearity). By PCA whitening, we reduce the dimensionality to 128.

\subsection{High-level feature modeling via sparse coding}
We use sparse coding to model high-level features that can train classifiers for event detection. 

\begin{figure} [t]
\centering
\begin{subfigure}[b]{0.25\textwidth}
\centering
\includegraphics[width=0.47\textwidth]{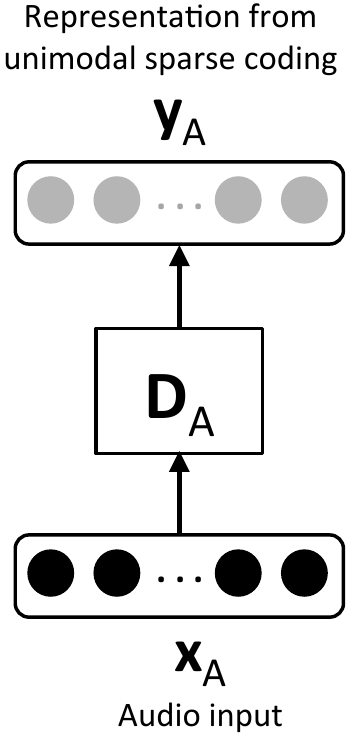}
\caption{Audio only}
\label{subfig:uni-audio}
\end{subfigure}
~
\begin{subfigure}[b]{0.25\textwidth}
\centering
\includegraphics[width=0.47\textwidth]{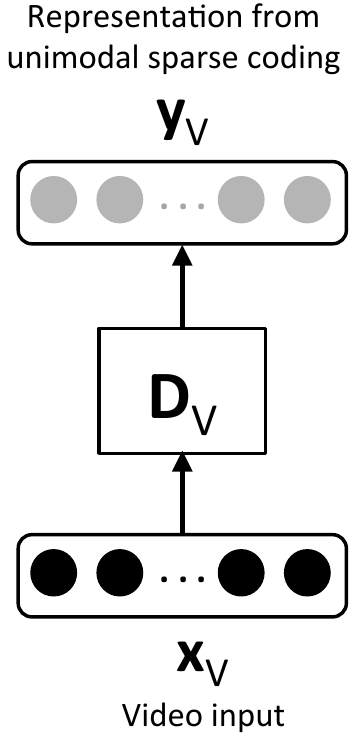}
\caption{Video only}
\label{subfig:uni-video}
\end{subfigure}
~
\begin{subfigure}[b]{0.4\textwidth}
\centering
\includegraphics[width=0.6\textwidth]{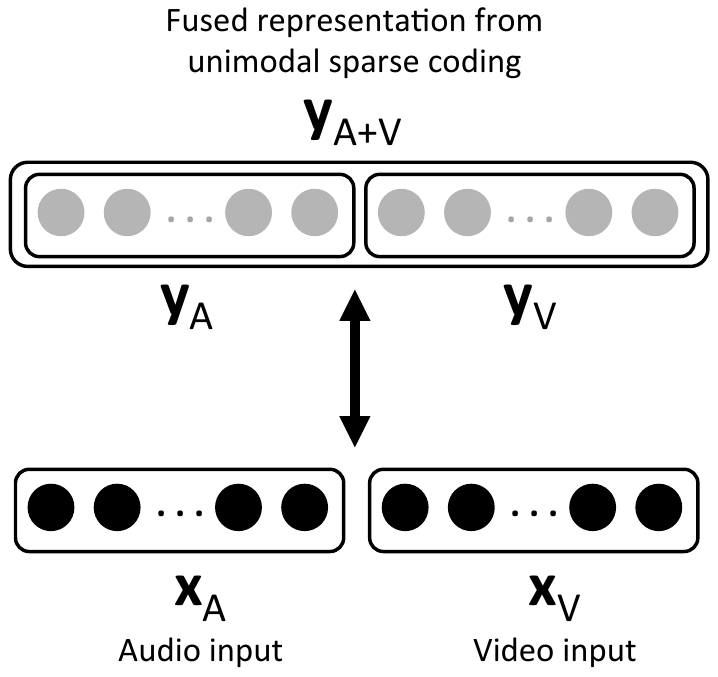}
\caption{Union of unimodal features}
\label{subfig:uni-fusion}
\end{subfigure}
\caption{Unimodal sparse coding and feature union}
\label{fig:uni}
\end{figure}

\textbf{Unimodal feature learning.} A straightforward approach for sparse coding with two heterogeneous data modalities is to learn a \emph{separate} dictionary of basis vectors for each modality. Figure~\ref{fig:uni} depicts unimodal sparse coding schemes. Recall the preprocessed audio and video input vectors $\mathbf{x}_\mathrm{A}$ and $\mathbf{x}_\mathrm{V}$. Audio-only sparse coding is done by \begin{align}
\min_{\mathbf{D}_\mathrm{A},\mathbf{y}_\mathrm{A}^{(i)}} \sum_{i=1}^{n_\mathrm{A}} \Arrowvert \mathbf{x}_\mathrm{A}^{(i)} - \mathbf{D}_\mathrm{A} \mathbf{y}_\mathrm{A}^{(i)} \Arrowvert^2_2 + \lambda \Arrowvert \mathbf{y}_\mathrm{A}^{(i)} \Arrowvert_1
\end{align} where we feed $n_\mathrm{A}$ unlabeled audio examples to simultaneously learn the unimodal dictionary $\mathbf{D}_\mathrm{A}$ and sparse codes $\mathbf{y}_\mathrm{A}^{(i)}$ under the sparsity regularization parameter $\lambda$. (We denote $\mathbf{x}_\mathrm{A}^{(i)}$ the $i$th training example for audio.) Similarly, using $n_\mathrm{V}$ unlabeled video examples, we learn \begin{align}
\min_{\mathbf{D}_\mathrm{V},\mathbf{y}_\mathrm{V}^{(i)}} \sum_{i=1}^{n_\mathrm{V}} \Arrowvert \mathbf{x}_\mathrm{V}^{(i)} - \mathbf{D}_\mathrm{V} \mathbf{y}_\mathrm{V}^{(i)} \Arrowvert^2_2 + \lambda \Arrowvert \mathbf{y}_\mathrm{V}^{(i)} \Arrowvert_1.
\end{align} We can form $\mathbf{y}_{\mathrm{A+V}} = [\mathbf{y}_\mathrm{A}~\mathbf{y}_\mathrm{V}]^\top$, a union of the audio and video feature vectors from unimodal sparse coding illustrated in Figure~\ref{subfig:uni-fusion}.

\begin{figure} [t]
\centering
\begin{subfigure}[b]{0.27\textwidth}
\centering
\includegraphics[width=0.8\textwidth]{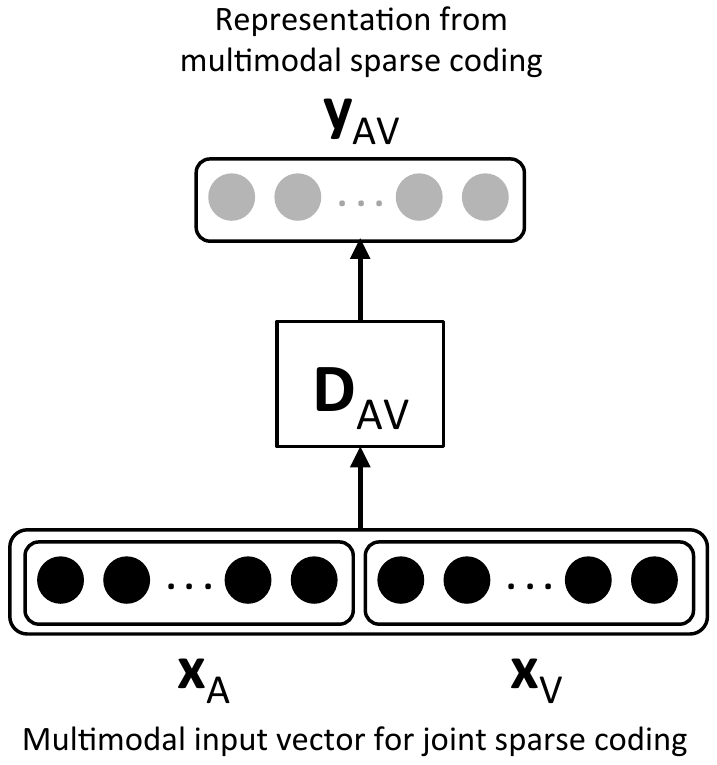}
\caption{$\mbox{\scriptsize Joint sparse coding}$}
\label{subfig:multi-av}
\end{subfigure}
~
\centering
\begin{subfigure}[b]{0.18\textwidth}
\centering
\includegraphics[width=0.65\textwidth]{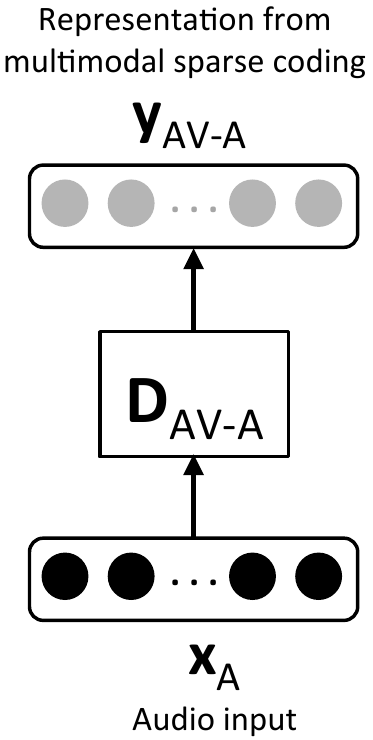}
\caption{$\mbox{\scriptsize Cross-modal audio}$}
\label{subfig:multi-audio}
\end{subfigure}
~
\begin{subfigure}[b]{0.18\textwidth}
\centering
\includegraphics[width=0.65\textwidth]{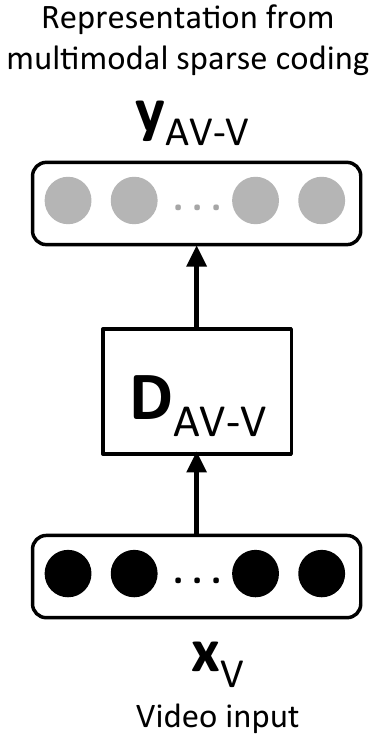}
\caption{$\mbox{\scriptsize Cross-modal video}$}
\label{subfig:multi-video}
\end{subfigure}
~
\begin{subfigure}[b]{0.27\textwidth}
\centering
\includegraphics[width=0.8\textwidth]{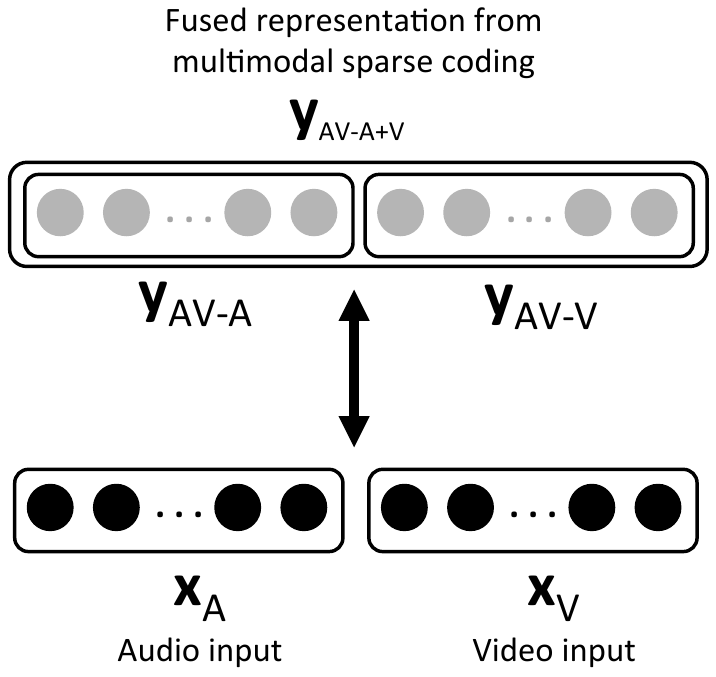}
\caption{$\mbox{\scriptsize Union of cross-modal features}$}
\label{subfig:multi-fusion}
\end{subfigure}
\caption{Multimodal sparse coding and feature formation possibilities}
\label{fig:multi}
\end{figure} 

\textbf{Multimodal feature learning.} The feature union $\mathbf{y}_{\mathrm{A+V}}$ encapsulates both audio and video sparse codes. However, the training is done in a parallel, unimodal fashion such that sparse coding dictionary for each modality is learned independently of the other. To remedy the lack of joint learning, we propose a multimodal sparse coding scheme described in Figure~\ref{subfig:multi-av}. We use the joint sparse coding technique used in image super-resolution \cite{jointsc} \begin{align} \label{eq:jointsc}
\min_{\mathbf{D}_\mathrm{AV},\mathbf{y}_\mathrm{AV}^{(i)}} \sum_{i=1}^{n} \Arrowvert \mathbf{x}_\mathrm{AV}^{(i)} - \mathbf{D}_\mathrm{AV} \mathbf{y}_\mathrm{AV}^{(i)} \Arrowvert^2_2 + \lambda' \Arrowvert \mathbf{y}_\mathrm{AV}^{(i)} \Arrowvert_1.
\end{align} Here, we feed the concatenated audio-video input vector $\mathbf{x}_\mathrm{AV}^{(i)} = [\frac{1}{\sqrt{N_\mathrm{A}}}\mathbf{x}_\mathrm{A}^{(i)}~\frac{1}{\sqrt{N_\mathrm{V}}}\mathbf{x}_\mathrm{V}^{(i)}]^\top$, where $N_\mathrm{A}$ and $N_\mathrm{V}$ are dimensionalities of $\mathbf{x}_\mathrm{A}$ and $\mathbf{x}_\mathrm{V}$, respectively. As an interesting property, we can decompose the jointly learned dictionary $\mathbf{D}_\mathrm{AV} = [\frac{1}{\sqrt{N_\mathrm{A}}}\mathbf{D}_\mathrm{AV-A}~\frac{1}{\sqrt{N_\mathrm{V}}}\mathbf{D}_\mathrm{AV-V}]^\top$ to perform the following audio-only and video-only sparse coding \begin{align} \label{eq:ja}
\min_{\mathbf{D}_\mathrm{AV-A},\mathbf{y}_\mathrm{AV-A}^{(i)}} \sum_{i=1}^{n_\mathrm{A}} \Arrowvert \mathbf{x}_\mathrm{AV-A}^{(i)} - \mathbf{D}_\mathrm{AV-A} \mathbf{y}_\mathrm{AV-A}^{(i)} \Arrowvert^2_2 + \lambda'' \Arrowvert \mathbf{y}_\mathrm{AV-A}^{(i)} \Arrowvert_1,\\ \label{eq:jv} \min_{\mathbf{D}_\mathrm{AV-V},\mathbf{y}_\mathrm{AV-V}^{(i)}} \sum_{i=1}^{n_\mathrm{V}} \Arrowvert \mathbf{x}_\mathrm{AV-V}^{(i)} - \mathbf{D}_\mathrm{AV-V} \mathbf{y}_\mathrm{AV-V}^{(i)} \Arrowvert^2_2 + \lambda'' \Arrowvert \mathbf{y}_\mathrm{AV-V}^{(i)} \Arrowvert_1.
\end{align} In principle, joint sparse coding via Eq.~(\ref{eq:jointsc}) combines the objectives of Eqs.~(\ref{eq:ja}) and (\ref{eq:jv}), forcing the sparse codes $\mathbf{y}_\mathrm{AV-A}^{(i)}$ and $\mathbf{y}_\mathrm{AV-V}^{(i)}$ to share the same representation. Note the relationship between the regularization parameters $\lambda' = (\frac{1}{N_\mathrm{A}} + \frac{1}{N_\mathrm{V}})\lambda''$. Ideally, we could have $\mathbf{y}_\mathrm{AV}^{(i)} = \mathbf{y}_\mathrm{AV-A}^{(i)} = \mathbf{y}_\mathrm{AV-V}^{(i)}$, although empirical values determined by the three different optimizations differ in reality. Feature formation possibilities on multimodal sparse coding are explained in Figure 3. 

%% file: eval.tex
\section{Evaluation}
\subsection{Dataset, task, and experiments}
We use the TRECVID MED 2014 dataset \cite{trecvidmed14} to evaluate our schemes. We consider the event detection and retrieval tasks using the 10Ex and 100Ex data scenarios, where 10Ex includes 10 multimedia examples per event, and 100 examples for 100Ex. There are 20 event classes (E021 to E040) with event names such as ``Bike trick," ``Dog show," and ``Marriage proposal." For evaluation, we compute classification accuracy and mean average precision (mAP) metrics according to the NIST standard on the following experiments:\vspace{-6pt} \begin{enumerate} \itemsep 0pt
\item Cross-validation on 10Ex;
\item 10Ex/100Ex (train with 10Ex and test on 100Ex).
\end{enumerate} \vspace{-4pt}

We use the number of basis vectors $K = 512$ same for all dictionaries $\mathbf{D}_\mathrm{A}$, $\mathbf{D}_\mathrm{V}$, and $\mathbf{D}_\mathrm{AV}$. We aggregate sparse codes around each keyframe of a training example by max pooling to form feature vectors for classification. We train linear, 1-vs-all SVM classifiers for each event whose hyper-parameters are determined by 5-fold cross-validation on 10Ex. We use the INRIA SPAMS (SPArse Modeling Software) \cite{spams}, VOICEBOX Speech Processing Toolkit \cite{voicebox}, MatConvNet \cite{matconvnet} to drive the pretrained deep CNN models, and LIBSVM \cite{libsvm}. 

\subsection{Other feature learning methods for comparison}
We consider other unsupervised methods to learn audio-video features for comparison. We evaluate the performance of Gaussian mixture model (GMM) and restricted Boltzmann machine (RBM) \cite{rbm} under similar unimodal and multimodal settings. For GMM, we use the expectation-maximization (EM) to fit the preprocessed input vectors $\mathbf{x}_\mathrm{A}$, $\mathbf{x}_\mathrm{V}$, $\mathbf{x}_\mathrm{AV}$ in 512 mixtures and form GMM supervectors \cite{gmmsv} as feature that contain posterior probabilities with respect to each Gaussian. The max-pooled GMM supervectors are applied to train linear SVMs. We adopt the shallow bimodal pretraining model by Ngiam \etal~\cite{mmdl} for RBM. Activations from the hidden layer of a size 512 are also max pooled before SVM. We have applied a target sparsity of 0.1 to both GMM and RBM.

\subsection{Results}
Table~\ref{tab:res-sc} presents the classification accuracy and mAP performance of unimodal and multimodal sparse coding schemes. For the 10Ex/100Ex experiment, we have used the best parameter setting from the 10Ex cross-validation to test 100Ex examples. In general, we observe that the union of audio and video feature vectors perform better than using only unimodal or cross-modal features. The union schemes perform better than the joint schemes. The union schemes, however, double feature dimensionality (\ie, from 512 to 1,024) since our union operation concatenates the two feature vectors. Joint feature vector is an economical way of combining both the audio and video features while keeping the same dimensionality as audio-only or video-only.

In Table~\ref{tab:others}, we report the mean accuracy and mAP for GMM and RBM under the union and joint feature learning schemes on the 10Ex/100Ex experiment. Our results show that sparse coding is better than GMM by 5--6\% in accuracy and 7--8\% in mAP. However, we find that the performance of RBM is on par with sparse coding. This leaves us a good next step to develop joint feature learning scheme for RBM. 

\begin{table}
\footnotesize
\centering
\caption{Mean accuracy and mAP performance of sparse coding schemes}
\begin{tabular}{|l|ccc|cccc|}
\hline
\multirow{3}{*}{} & \multicolumn{3}{c|}{Unimodal} & \multicolumn{4}{c|}{Multimodal}\\ 
& Audio-only & Video-only & Union & Audio & Video & Joint & Union \\ 
& {\scriptsize (Fig.~\ref{subfig:uni-audio})} & {\scriptsize (Fig.~\ref{subfig:uni-video})} & {\scriptsize (Fig.~\ref{subfig:uni-fusion})} & {\scriptsize (Fig.~\ref{subfig:multi-audio})} & {\scriptsize (Fig.~\ref{subfig:multi-video})} & {\scriptsize (Fig.~\ref{subfig:multi-av})} & {\scriptsize (Fig.~\ref{subfig:multi-fusion})} \\ \hline \hline
\begin{tabular}[c]{@{}l@{}} Mean accuracy\\ (cross-val. 10Ex)\end{tabular} & 69\% & 86\% & \textbf{89\%} & 75\% & 87\% & 90\% & \textbf{91\%} \\ \hline
\begin{tabular}[c]{@{}l@{}} mAP\\ (cross-val. 10Ex)\end{tabular} & 20.0\% & 28.1\% & \textbf{34.8\%} & 27.4\% & 33.1\% & 35.3\% & \textbf{37.9\%}\\ \hline \hline
\begin{tabular}[c]{@{}l@{}} Mean accuracy\\ (10Ex/100Ex)\end{tabular} & 56\% & 64\% & \textbf{71\%} & 58\% & 67\% & 71\% & \textbf{74\%} \\ \hline
\begin{tabular}[c]{@{}l@{}} mAP\\ (10Ex/100Ex)\end{tabular} & 17.3\% & 28.9\% & \textbf{30.5\%} & 23.6\% & 28.0\% & 28.4\% & \textbf{33.2\%}\\ \hline
\end{tabular}
\label{tab:res-sc}
\end{table}

\begin{table}
\footnotesize
\centering
\caption{Mean accuracy and mAP performance for GMM and RBM on 10Ex/100Ex}
\begin{tabular}{|l|c|c|}
\hline
Feature learning schemes  & Mean accuracy & mAP \\ \hline
\begin{tabular}[c]{@{}l@{}} Union of unimodal GMM features\\ (Figure~\ref{subfig:uni-fusion})\end{tabular}  & 66\% & 23.5\% \\ \hline
\begin{tabular}[c]{@{}l@{}} Multimodal joint GMM feature\\ (Figure~\ref{subfig:multi-av})\end{tabular}    & 68\% & 25.2\% \\ \hline \hline
\begin{tabular}[c]{@{}l@{}} Union of unimodal RBM features\\ (Figure~\ref{subfig:uni-fusion})\end{tabular}  & 70\% & 30.1\% \\ \hline
\begin{tabular}[c]{@{}l@{}} Multimodal joint RBM feature\\ (Figure~\ref{subfig:multi-av})\end{tabular} & 72\% & 31.3\% \\ \hline
\end{tabular}
\label{tab:others}
\end{table}

%% file: conc.tex
\section{Conclusion}
We have presented multimodal sparse coding for MED. Our approach can build joint sparse feature vectors learned from different modalities and scale to file-level descriptors suitable for training classifiers in a MED system. Using the TRECVID MED 2014 dataset, we have empirically validated our approach and achieved promising performance measured in accuracy and precision metrics recommended by the NIST standard. Our future work includes an integration with more features, training for larger event coverage, and finetuning.